\documentclass{article}
\usepackage[utf8]{inputenc}
\usepackage[backend=biber,style=numeric,sorting=nyt,maxbibnames=99]{biblatex}
\usepackage{graphics}
\usepackage{adjustbox}
\usepackage{amsfonts}
\usepackage{graphicx}
\usepackage{changepage}
\usepackage{amsmath,bm}
\usepackage{afterpage}
\usepackage{lscape}
\usepackage{caption}
\usepackage{subcaption}
\usepackage{float}
\usepackage{algorithm}
\usepackage{algorithmicx}
\usepackage{algpseudocode}
\usepackage{capt-of}
\usepackage{arxiv}
\graphicspath{ {./Images/} }
\begin{document}

\title{NICE: An Algorithm for Nearest Instance Counterfactual Explanations}

\author{Dieter Brughmans \and Pieter Leyman \and David Martens 
}


\maketitle

\begin{abstract}
In this paper we suggest NICE: a new algorithm to generate counterfactual explanations for heterogeneous tabular data. The design of our algorithm specifically takes into account algorithmic requirements that often emerge in real-life deployments: (1) the ability to provide an explanation for \textit{all} predictions, (2) being able to handle any classification model (also non-differentiable ones), and (3) being efficient in run time. More specifically, our approach exploits information from a nearest unlike neighbour to speed up the search process, by iteratively introducing feature values from this neighbour in the instance to be explained. We propose four versions of NICE, one without optimization and, three which optimize the explanations for one of the following properties: sparsity, proximity or plausibility. An extensive empirical comparison on 40 datasets shows that our algorithm outperforms the current state-of-the-art in terms of these criteria. Our analyses show a trade-off between on the one hand plausibility and on the other hand proximity or sparsity, with our different optimization methods offering users the choice to select the types of counterfactuals that they prefer. An open-source implementation of NICE can be found at https://github.com/ADMAntwerp/NICE.

\keywords{XAI \and Counterfactual Explanations \and Machine Learning}

\end{abstract}

\section{Introduction}\label{intro}

\subsection{The need for explainability}

In the past decade, machine learning (ML) models have been successfully deployed in many high-stakes decision making domains such as credit scoring~\cite{lessmann2015benchmarking}, fraud detection~\cite{ngai2011application,whitrow2009transaction,callahan2017machine}, and clinical healthcare~\cite{callahan2017machine,huang2015mining}. However, due to the complexity of the models or high-dimensionality of the underlying data, for many models high performance has come at a cost of explainability~\cite{SEDC,wachter2018,ramon2020comparison}. The inability to explain automated decisions that impact individuals undermines the trust between data subject and data controllers~\cite{wachter2018}. Post-hoc explanation methods such as counterfactual explanations aim to reinstall this trust while keeping the performance of the decision mechanism~\cite{wachter2018}. Several types of legislation have also pushed on providing explanations for algorithmic decision-making. One example is the Fair Credit Reporting Act in the United States~\cite{FCRA}. It requires data controllers to provide specific reasons that negatively influence a data subject's credit score. Counterfactual explanations are a great fit here as they provide a set of minimum features required to change the predicted outcome. Another example comes from the General Data Protection Regulation (GDPR) in the European Union. Article 14 states that data subjects have the right to obtain meaningful information about the logic involved in automated decision making~\cite{GDPR}. 

Current classification models have a high complexity and many parameters. As a consequence, explaining the inner working of such a model will not be meaningful to a data subject. Counterfactual explanations, on the contrary, highlight a set of input features that, when changed, alter the predicted decision~\cite{SEDC}. These input features are a lot more understandable to humans as the form of counterfactual explanations has deep foundations in philosophy~\cite{kment2006counterfactuals,lewis2013counterfactuals,ruben2015explaining} and social sciences~\cite{MILLER20191}; for it is similar to how a person thinks about a decision by asking the question: what could I have changed to achieve a different outcome? Additionally, counterfactual explanations allow data controllers to explain instances without disclosing any trade secrets or private data~\cite{barocas2020hidden}.

It is clear that in theory counterfactual explanations fit the legislative requirements and have the ability to make black-box ML models transparent and accountable. Spurred by these benefits, in recent years many counterfactual algorithms have been developed for tabular data (see e.g. the overviews by~\cite{verma2020counterfactual,karimi2020survey}). However, most of them focus on generating counterfactuals without taking into account the algorithmic requirements in deployment. Consider for example custom fraud detection. In a country such as Belgium, custom administration processes around 9.5 declarations every second~\cite{vanhoeyveld2020value}. Each of these cases have the potential for different forms of fraud such as illegal drug traffic, importation of counterfeit goods, valuation fraud, smuggling, product misclassification and the manipulation of the origin of goods. Predictive algorithms are used in this context to identify high-risk targets which are further investigated by custom officers \cite{vanhoeyveld2020value,digiampietri2008uses}. Counterfactual explanations can be of great value here to improve collaboration. The set of features in this explanation can clarify which form of fraud might be committed, or what the main reasons for predicted fraud are (e.g. country risk, article code, weight and so on), thereby speeding up further investigations. Custom officers check many high-risk cases each minute, so explanation algorithms will have to match this speed to make them useful. This computational efficiency requirement also guarantees that these algorithms can be easily scaled without the need of excessive infrastructure. An additional requirement is that \textit{all} observations can be explained in this domain. The absence of an explanation might give the (potentially wrong) impression that the predictive model is not certain about its prediction, undermining the trust in the predictive model and making it difficult for custom officers to act upon the output.

If we return to the example of credit scoring in the US, we notice the same requirements. In this case explanations for negative decisions are required by law, rendering counterfactual algorithms that cannot explain all of these credit rejection predictions, useless. Also computational inefficiency is costly here. In credit scoring, the data subjects are potential customers. Imagine a consumer applying for a loan at a bank. A number of variables are asked, such as income, profession, etc. to assess the credit risk. The classification model, which is efficient by design as well, as not to have the consumer wait for minutes or hours to get a decision, will provide a decision swiftly. In case the application is rejected, it is just as important to have an efficient explanation algorithm to come up with a motivation for the rejection. Having the consumer just sit there and wait is arguably unacceptable or at least bad business practice. 

\begin{figure*}
  \includegraphics[width=1\textwidth]{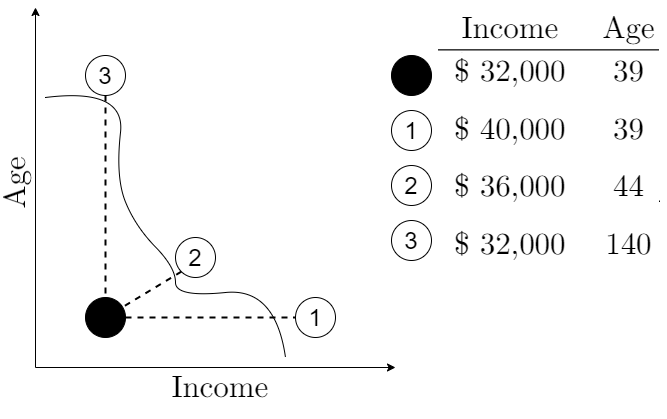}
\caption{Example of counterfactual explanations for loan approval.}
\label{CFexample}       
\end{figure*}

In Figure~\ref{CFexample} we show a simplified example of a counterfactual explanation in the domain of credit scoring. In this example the loans of individuals will be approved or denied based on their income and age. The graph shows a two-dimensional feature space. The black line represents a classification model that splits the feature space into two areas: The loans of individuals who land on the right will be approved, while the individuals who land on the right will be denied. As an example, consider a person, 39 years old and with an income of \$32,000, who applies for a loan. This person is represented by the black dot and the classification model denies the loan. If this person receives a raise, increasing her income by \$8,000, she would land on white dot number 1 in the feature space where her loan will be approved. This \$8000 increase in income is an example of a counterfactual explanation. It is the change that has to be made to an observation in order to change its predicted class. This raise would result in the person being 39 years old and earning \$40,000, which is called the counterfactual instance.

\subsection{Research objectives}

ML in general is a fast evolving field where models vary over applications and time. Current state-of-the-art classification models, might be outperformed in a few years. To ensure that counterfactual algorithms remain useful when classification models change, and to give data controllers full freedom over the choice of these models, there is a preference for model-agnostic explanation algorithms. This implies that the classification model is used simply as an output generating machine, based on provided input. We have thus identified three important algorithmic requirements:
\begin{itemize}
    \item Perfect coverage, which means we want the counterfactual algorithm to be able to provide an explanation for each prediction.
    \item Required model access, which preferably limits itself to the inputs and outputs of the models, making the algorithm model-agnostic.
    \item Computational efficiency.
\end{itemize}

In this paper, we propose Nearest Instance Counterfactual Explanations (NICE), a new algorithm to find counterfactual explanations for tabular data with both numerical and categorical variables (summarized hereafter as heterogeneous tabular data). This type of data is widely used in ML applications where individuals are impacted, such as credit scoring, clinical healthcare, recruitment or fraud detection. Explanations are extremely important in this context as a wrong decision can have dire consequences~\cite{doshi2017towards}.

Our contribution to the XAI literature, and more specifically to the counterfactual literature, amounts to the introduction of a new algorithm, which allows for \textbf{simultaneously} achieving the following:
\begin{enumerate}
    \item \textbf{100\% coverage:} NICE always finds a counterfactual and hence can always find an explanation for a given instance and a given classification model.
    \item \textbf{Model-agnostic:} NICE works for any kind of classification model and does not require any kind of information regarding the internal workings of the classification model. 
    \item \textbf{Fast counterfactual generation:} NICE is fast, especially in light of the required 100\% coverage (see Section \ref{exp} for the computational experiments).
\end{enumerate}

Whereas other algorithms permit one or two of the above (see Section \ref{bench}), NICE guarantees all of these simultaneously (see Section \ref{meth}). NICE furthermore employs existing counterfactual properties (i.e., sparsity, proximity and plausibility), and is built in such a way that properties can be integrated as objectives in a modular way, depending on the users' requirements.


The remainder of this paper is structured as follows. We discuss related work in Section \ref{lit} and explicitly consider potential benchmark algorithms for NICE. We formally introduce NICE itself in Section \ref{meth} and show detailed computational results in Section \ref{exp}. Finally, we present our conclusions along with future research avenues in Section \ref{concl}.

\section{\label{lit}Related work}
Explainable AI (XAI), of which the goal is to explain individual model predictions resulting from black-box models, has seen tremendous growth over the past decade. Established and recent explanation algorithms include LIME \cite{ribeiro2016}, SHAP \cite{lundberg2017}, CAVs \cite{kim2018}, Anchors \cite{ribeiro2018}, ProtoPNet \cite{chen2019} and SpRAy \cite{lapuschkin2019}. Since in this research we focus on counterfactuals as technique for explainability, we mainly restrict ourselves to counterfactual research in the following subsections. For in-depth overviews on XAI and its existing algorithms, we refer to Molnar \cite{molnar2019} and Barredo et al. \cite{barredo2020}.

In Section \ref{cf} we give an overview of the recent and relevant counterfactual literature in general, whereas in Section \ref{bench} we go into detail about possible benchmark algorithms. In Section \ref{cfmetrics} we discuss properties of counterfactuals, and finally we touch upon user interpretability in Section \ref{user}.

\subsection{\label{cf}Counterfactuals}
Martens \& Provost~\cite{SEDC} have been argued to be the first to formally introduce a counterfactual evidence method (see overview by Karimi et al. \cite{karimi2020survey}), which they call SEDC\footnote{Pronounced as ``Set See''.}. Martens \& Provost applied SEDC to textual data, while in subsequent work it has been employed for behavioral data~\cite{ramon2020comparison}, image data~\cite{vermeire2022explainable} and finally tabular data~\cite{fernandez2020} as well. Other interesting forerunners of counterfactuals are the case-based reasoning approaches by Nugent \& Cunningham \cite{nugent2005} and Nugent et al. \cite{nugent2009}, which provide insight into the impact of feature value changes on model predictions, although they focus on actual past cases while counterfactuals discuss required changes in the current situation. Given the similarities between case-based reasoning and counterfactuals as explanation methods, we discuss the case-based reasoning approach by Keane \& Smyth \cite{keane2020good}, specifically applied to counterfactuals, in more detail below, and also compare our results with theirs (see Section \ref{exp}).

Wachter et al.~\cite{wachter2018} stated the problem as a loss function to be optimized. With this approach it is easy to impose desired properties on the explanations by adding extra terms to this loss function. Mothilal et al.~\cite{mothilal2020explaining} added a diversity parameter to generate multiple counterfactual explanations for each observation. Others implemented an auto-encoder (AE)~\cite{CFproto,dhurandhar2018explanations} or prototype~\cite{CFproto} loss, resulting in explanations closer to the data manifold. However, for \textit{heterogeneous} tabular data, these loss functions face some challenges. First of all, they have difficulties handling nominal variables. Some solve this problem by one-hot encoding the variables and adding an extra loss term to enforce a correct encoding~\cite{mothilal2020explaining,joshi2019towards}. Others map the features into an ordinal vector space~\cite{dhurandhar2019model,CFproto}. A more substantial problem is that these loss functions can only be solved efficiently when gradients are available, which is only the case for differentiable models. To date, the best performing models for tabular data often include tree-based ensembles~\cite{olson2017pmlb,lessmann2015benchmarking}. For these non-differentiable models, the gradients have to be calculated numerically which causes a computational bottleneck. Van Looveren and Klaise~\cite{CFproto} claim to have reduced this bottleneck by adding the distance to the training data to the loss function, causing the optimization to converge faster.

Delaney et al. \cite{delaney2020} discuss counterfactuals in case of time series data and develop Native Guide as solution approach, which adapts existing counterfactual instances. The authors provide details of characteristics of good counterfactuals, and use these in their algorithm. The results show the superiority of Native Guide compared to existing techniques for time series data.

Multi-objective counterfactual explanations are for the first time discussed by Dandl et al. \cite{dandl2020multi}, whose approach allows for a better trade-off between different objectives and is based on the multi-objective optimization literature. An adjusted version of the nondominated sorting genetic algorithm (NSGA-II) is used to solve the multi-objective problem.

Finally, Mothilal et al. \cite{mothilal2021} seek to combine feature importance analysis with counterfactual generation, since both types of approaches may highlight different features. The authors study the necessity and sufficiency of features and find that not only the top features reported by feature importance methods can prove to be meaningful for explanations.

\subsection{\label{bench}Potential benchmark algorithms}

We give an overview of counterfactual algorithms worth considering to benchmark NICE (experiments discussed in detail in Section \ref{exp}) in Table \ref{tabalgs}. For a recent overview on and classification of the counterfactual literature in general, we refer to Karimi et al. \cite{karimi2020survey}, whereas de Oliveira \& Martens \cite{deoliveira2021} explicitly perform a benchmarking study for ten counterfactual algorithms on tabular data.

Because we focus on a new optimization approach, the algorithms in Table \ref{tabalgs} have been grouped into ``families'' according to the type of algorithm used, that is, gradient descent, neighbor-based, genetic algorithm and SAT solver. Combined with the more general survey on counterfactual algorithms by Karimi et al. \cite{karimi2020survey}, we can conclude that gradient descent and neighbour-based heuristics are the most prevalent types of algorithms currently used. The overview in Table \ref{tabalgs} furthermore takes three properties from the literature used by NICE into account as objectives (sparsity, proximity and plausibility, see Sections \ref{cfmetrics} and \ref{sec:Reward Functions}), along with custom specifications 100\% coverage guarantee and compatibility with heterogeneous tabular data (numerical and categorical features). For algorithm specifics, aside from the type of algorithm, we include the model-agnosticism of the algorithm and whether it requires access to the classification model. Finally, we show whether Python (compatible) code is publicly available or not. In the following paragraphs, we briefly discuss each of the articles in Table \ref{tabalgs}. Note that some of these articles have already been touched upon in earlier paragraphs, in order to better situate our own work. 

\begin{table}
	\footnotesize
	\begin{center}
		\begin{tabular}{l|ccc|cc|ccc}
		    & \multicolumn{3}{c|}{\textbf{Objectives}} & \multicolumn{2}{c|}{\textbf{Custom}} & \multicolumn{3}{c}{\textbf{Algorithm}}\\ 
			\textbf{Name} & \textbf{Spars.} & \textbf{Prox.} & \textbf{Plaus.} & \textbf{Cover.} & \textbf{HT}&\textbf{Type} & \textbf{Model-}&\textbf{Code}\\
			& & & & &\textbf{data} & & \textbf{agn.} &\\
			\hline
			DiCE \cite{mothilal2020explaining} & & x & & & x & GD & Y* & Y \\
			CFproto \cite{CFproto} & x & x& x & & x & GD & Y* & Y \\
			\hline
			Growing Spheres \cite{laugel2017} & x & x & & & & NB & Y & Y \\
			WIT \cite{wexler2019if} & & x & x& & x & NB & Y &  N*\\
			CBR \cite{keane2020good} & x& &x& & x& NB & Y & N* \\
			SEDC \cite{fernandez2020} & x & & & & x& NB & Y &  Y \\
			NICE (this paper)& x & x & x & x &x & NB & Y &  Y \\
			\hline
			GeCo \cite{schleich2021} & x & x &x & & x & GA & Y &  Y \\
			\hline
			MACE \cite{karimi2020model} & x &x &x & x & x& SAT & Y & Y$^\dagger$ \\
		\end{tabular}
	\end{center}
	\caption{Overview counterfactual algorithms for tabular data \textbf{potentially relevant for comparison}. Objectives present in a paper are marked with an ``x'' in the corresponding column for sparsity (Spars.), proximity (Prox.) and plausibility (Plaus.), as are the coverage guarantee (Cover.) and the use of heterogeneous tabular data (HT data). Related to the algorithms themselves, the type (GD for gradient descent, NB for neighbour-based, GA for genetic algorithm and SAT for SAT solver) is included, along with whether it is model-agnostic (yes or no, an asterisk means model access is an optional feature) and whether the code is publicly available and usable in Python without further assumptions (yes or no, an asterisk * means we implemented the algorithm ourselves, a dagger $^\dagger$ means specific code assumptions are made).}
	\label{tabalgs}
\end{table}

In terms of \textbf{gradient descent} algorithms, DiCE (Diverse Counterfactual Explanations) by Mothilal et al. \cite{mothilal2020explaining} allows for the generation of diverse counterfactuals for the same instance by combining multiple objectives into a single loss function. The approach returns multiple counterfactuals for a given instance, in order to present a user with several diverse counterfactuals. Another gradient descent algorithm is CFproto\footnote{This name comes from the GitHub repository of Van Looveren \& Klaise \cite{CFproto}, in which CFproto is the name used for the .py file which contains their algorithm.} by Van Looveren \& Klaise \cite{CFproto}. Unlike some of its gradient-based peers, it has found a way to work with categorical data in the loss function and claims to be faster in optimizing it. The purpose of this algorithm is very similar to that of NICE (\texttt{plaus}). It tries to find a balance between close and plausible explanations and uses the training data to achieve this \cite{CFproto}.

The following algorithms all fall under the \textbf{neighbour-based} family. Growing Spheres \cite{laugel2017} focuses on finding the minimal required changes to an instance in order to obtain a counterfactual, without relying on existing data. Growing Spheres, however, is only compatible with numerical features and does not take categorical features into account.\cite{deoliveira2021} The What-if Tool (WIT)~\cite{wexler2019if} is an interactive tool which selects the nearest instance from the training set that is classified in a different class. Besides two small differences, this method is exactly the same as NICE (\texttt{none}), a basic version of NICE (see Section \ref{niceSection}). First, WIT selects a counterfactual instance from the complete training set and not only the correctly classified ones. Second, the distance metric is slightly different: for numerical features it standardizes the differences with the standard deviation and not the range of the feature values in the training set. In spite of these two small differences, WIT would make a good benchmark for NICE. Just like NICE and WIT, the Case-Based Reasoning system (CBR) for counterfactual explanations \cite{keane2020good} uses nearest instances to find counterfactuals. The difference is the optimization method, which limits the search to explanations that have a maximum sparsity of two features. Furthermore, this method also does not guarantee that an explanation will be found (coverage), but it could still be worthwhile as a point of comparison. The next algorithm is SEDC for tabular data \cite{fernandez2020}. The optimization method is very similar to NICE under sparsity. The difference between both algorithms is their search space. Whereas NICE replaces feature values with those of the nearest instance, SEDC replaces them with the respective mean or mode of each feature.

A \textbf{genetic algorithm} for counterfactual generation (GeCo), which focuses on finding counterfactuals with minimal changes, is proposed by Schleich et al. \cite{schleich2021}. The algorithm prefers explanations with as few as possible changes in feature values, and furthermore allows for the comparatively fast generation of counterfactuals.

Karimi et al. \cite{karimi2020model} build on formal logic and solve a sequence of \textbf{satisfiability problems (SAT)}, as part of their counterfactual approach called Model-Agnostic Counterfactual Explanations (MACE). Their results show that MACE has 100\% coverage, i.e., the algorithm can always find a counterfactual, though its computation times are negatively affected by the use of an SAT solver. The open-source Python implementation of MACE\footnote{https://github.com/amirhk/mace} also sets some specific constraints on the encoding of categorical variables, which are not compatible with other models in our experiments. For these reasons, we decided not to employ MACE as a benchmark algorithm. 

In order to allow for a proper and fair comparison with NICE, in the sense that benchmarks should have similar characteristics as our approach, we focus on algorithms with the following characteristics: (1) model-agnostic and no model access, (2) use of heterogeneous tabular data, and (3) code available and usable without further assumptions. Though both WIT and CBR, two algorithms very similar to NICE, have no Python code publicly available, we decided to implement both algorithms ourselves (code available online in our GitHub repository). Both CBR and WIT are quite similar to our approach, which warrants a comparison, and are furthermore straightforward enough to implement them ourselves. Combined with the three conditions above, the remaining algorithms are those in Table \ref{tabalgsrest}, which means that we have six benchmark algorithms for NICE. From the table, we conclude that in the selected benchmark algorithms there is variation in both the objectives and the algorithm type, while NICE and WIT are the only ones that have 100\% coverage.

\begin{table}
	\footnotesize
	\begin{center}
		\begin{tabular}{l|ccc|c|c}
		    & \multicolumn{3}{c|}{\textbf{Objectives}} & \multicolumn{1}{c|}{\textbf{Custom}} & \multicolumn{1}{c}{\textbf{Algorithm}}\\ 
			\textbf{Name} & \textbf{Spars.} & \textbf{Prox.} & \textbf{Plaus.} & \textbf{Cover.} & \textbf{Type} \\
			\hline
			DiCE \cite{mothilal2020explaining} & & x & & & GD  \\
			CFproto \cite{CFproto} & x & x& x & & GD  \\
			\hline
			WIT \cite{wexler2019if} & & x &x& & NB  \\
			CBR \cite{keane2020good} & x& &x& &  NB  \\
			SEDC \cite{fernandez2020} & x & & & &  NB  \\
			NICE (this paper)& x & x & x & x  & NB  \\
			\hline
			GeCo \cite{schleich2021} & x & x &x & & GA \\
		\end{tabular}
	\end{center}
	\caption{Overview counterfactual algorithms for tabular data \textbf{used for comparison}. Objectives present in a paper are marked with an ``x'' in the corresponding column for sparsity (Spars.), proximity (Prox.) and plausibility (Plaus.), as is the coverage guarantee (Cover.). Related to the algorithms themselves, the type (GD for gradient descent, NB for neighbour-based, GA for genetic algorithm) is included.}
	\label{tabalgsrest}
\end{table} 

\subsection{\label{cfmetrics}Counterfactual properties}
Previous research has pointed out several important properties of counterfactual explanations \cite{karimi2020survey,verma2020counterfactual}. In this work, we focus on sparsity, proximity and plausibility.

A commonly used property is \textbf{sparsity} \cite{karimi2020model,dandl2020multi,laugel2018comparison} which refers to the number of features in an explanation. It is often claimed that sparser explanations are better as they are less complex. This statement stems from psychological research which finds that people can only process five to nine pieces of information at once \cite{miller1956magical,medin1987,edwards2019,forster2020,forster2021}. Especially when working with high dimensional features spaces, a sparsity constraint is useful to ensure explanations remain comprehensible for humans. In Figure \ref{CFexample}, counterfactual 1 is sparser than counterfactual 2. If loan applicants only want to change their income and do not want to wait until they get older, counterfactual 1 is probably a better explanation for them. Furthermore, consider that sparsity is actually equal to the L0 distance between the instance to explain and the counterfactual instance. A critique of sparsity is that it is a very rough measure of distance. For example, when providing explanations in a credit approval context, it is straightforward to get an explanation that suggests a \$200 raise is better than one which suggests a \$10,000 one. However, in terms of sparsity, both explanations are equal. To counter this, previous research has used alternative distance metrics such as L1 distance\cite{wexler2019if}, L2 distance \cite{mothilal2020explaining} or ABDM \cite{CFproto}. In this paper, we use \textbf{proximity} as an overlapping term for all these distance metrics. Therefore, sparsity can actually be seen as a specific example of proximity. However, we keep them as separate properties in this article because much of the previous literature specifically focuses on sparsity.

Pawelczyk et al. \cite{pawelczyk2020counterfactual} pointed out that spars or near counterfactual explanations may be vulnerable to classification model changes over time. The counterfactual instance possibly ends up in an area far from the data manifold, where these models predict with high uncertainty. When a different classification model is trained on the same data, the previous explanation might no longer be valid. In our loan approval example, this would correspond to a case where applicants are told to raise their income by \$8000. However, when they return to the bank, another model has been put into production and their loan request is again rejected. Such occurrences would diminish confidence in counterfactual explanations. In the rest of this paper, we call this concept cross-model robustness.

One property of counterfactual explanations that can avoid the problem of cross-model robustness is \textbf{plausibility} \cite{pawelczyk2020counterfactual}. It measures the closeness of the counterfactual instance to the data manifold. If an explanation in the loan approval example, such as counterfactual 3, suggests that applicants wait until they are 140 years old, this explanation is clearly not plausible, as it lies far outside of the data manifold. Compared to the previous two properties, plausibility is more conceptual and cannot be measured directly. Proxies that have been used to measure plausibility are: the distance to the k-nearest neighbors from the training data \cite{dandl2020multi}, the local outlier factor \cite{kanamori2020dace}, IM1 \cite{CFproto},IM2\cite{CFproto} and the reconstruction error of an AE trained on the training data \cite{mahajan2019preserving,CFproto}. Some studies have shown that there is an inherent trade-off between sparsity and plausibility \cite{dandl2020multi,CFproto}.

\subsection{\label{user}User interpretability}
We should not ignore the often context dependent user preferences related to XAI methods. Whereas we mainly discuss user interpretability here, overviews of general and often cross-disciplinary XAI requirements can be found in Miller \cite{MILLER20191} and in Langer et al. \cite{langer2021}.

As stated by among others Keane et al. \cite{keane2021}, a major deficit of counterfactual research is the small number of user studies, in order to gain insights into the type of counterfactuals (or explanations in general) users may prefer. As discussed before, the claimed preference for sparse explanations stems from psychological research \cite{miller1956magical,medin1987,edwards2019,forster2020,forster2021}.

In spite of the above, studies related to XAI have shown that user preferences on interpretability are context and requirements dependent, and that users may even prefer more complex explanations \cite{ramon2021,furnkranz2020cognitive}. Other recent examples of user studies are Byrne \cite{byrne2019} and Dodge et al. \cite{dodge2019}, with the latter explicitly concluding that there is no one-size-fits-all approach to explaining, but that the usefulness of explanations depends on user profiles and expertise. Alternatively, Weld \& Bansal \cite{weld2019} argue in favor of interactive explanation systems, as users may have follow-up questions and more detailed concerns once an (initial) explanation has been provided.

Fern\'{a}ndez-Lor\'{i}a et al. \cite{fernandez2020} develop an algorithm to detect the most useful counterfactual explanations based on context and compare their approach with feature importance techniques in three case studies. The authors take adjustable feature weights into account, allowing their approach to suggest different kinds of counterfactuals depending on user requirements, and conclude that features with a large impact on model prediction may not necessarily be good for explaining individual decisions. In this way, their method can be seen as a necessary middle ground between the development of the algorithm on the one hand and user studies on the other. 

\section{\label{meth}Methodology}
In this section, we propose NICE: a nearest unlike neighbor-based approach to generate counterfactual explanations. NICE applies a depth-first heuristic approach that guarantees to always find a counterfactual, which is a hybrid between the nearest unlike neighbour and the instance to explain. Consider that this nearest unlike neighbor is not only a counterfactual due to the class change, but also an existing instance from the dataset used.

In the remainder of this section, we first explain the search process of our algorithm step by step. Afterward, we provide more information on the different reward functions used to guide this process. 

\subsection{\label{niceSection}NICE overview}
Assume an $m$-dimensional feature space $X \subset \mathbb{R}^m$  consisting of both categorical and numerical features, a feature vector $ x\in X$ with a corresponding label denoted as $y \in Y = \{-1,1\}$ and a trained classification model $f$ which maps $\mathbb{R}^m$ in the class score vector such that $f(x) \in [-1,1]$ which leads to a predicted class $\hat{y}$. A counterfactual instance $x_c$  for $x_0$ minimizes the distance $d(x_0,x_c)$ under the condition that $\hat{y_0} \neq \hat{y_c}$. 

Our algorithm is very flexible in its distance metric as it does not need the categorical variables to be mapped in an ordinal vector. In this paper, we choose the Heterogeneous Euclidean Overlap Method (HEOM) as a distance metric \cite{wilson1997improved}. For each feature (F), the distance is calculated according to Formula (\ref{HEOM}), while the total distance is simply the L1-norm of all feature distances. Furthermore,this metric guarantees that the contribution of each feature to the total distance is between 0 and 1.

\begin{equation}
d_F(a,b)=
\begin{cases}
  1 \text{      if  } a \neq b \text{  for categorical F}\\
  0 \text{      if  } a = b \text{  for categorical F}\\
  \frac{\mid a-b \mid}{range(F)} \text{  for numerical F }
\end{cases}
\label{HEOM}
\end{equation}

Algorithm \ref{nice_alg} gives an overview of NICE and its constituting steps. The input of NICE is an instance $x_0$ for which we want to find a counterfactual instance, whereas the output $x_c$ is such a counterfactual instance. Consider that upon termination of NICE, $x_c$ will always remain a combination of $x_0$ and $x_n$, which significantly reduces our search space and consequently the run time of NICE. In Step 1 (lines 4-6), a current hybrid instance $x_c$ is created between $x_0$ and $x_n$ to keep track of the changes made to $x_0$ as the algorithm progresses, and a counter $i$ is initialized for Step 3. In Step 2 (lines 8-9), NICE finds the nearest unlike neighbor $x_n$ for $x_0$, for which $\hat{y}_0 \neq \hat{y}_{n}$ and $y_{n} = \hat{y}_{n}$, and subsequently identifies the non-overlapping features. This results in a list of features (\textit{featureList}), which can be changed to make $x_c$ more similar to $x_n$. This approach is similar to that of Delaney et al. \cite{delaney2020}, with the major difference that the latter use time series instead of tabular data.

Before moving on with step 3, note that $x_{n}$ from Algorithm \ref{nice_alg} can already be used as a counterfactual instance and has some desirable properties. First, it is an actual instance, which makes it by definition plausible. In addition, the second condition ($y_{n} = \hat{y}_{n}$) implies that the observation is correctly classified by $f$. Therefore, $x_{n}$ corresponds to an area in $\mathbb{R}^m$ where the predictions of $f$ are arguably more justified. If the classification model $f$ is replaced by a different one $h$, trained on the same data, there would be a higher probability that $x_{n}$ is also a counterfactual instance \cite{pawelczyk2020counterfactual}. We will refer to this version without further optimization as NICE(\texttt{none}) in the remainder of this work.

Step 3 of Algorithm \ref{nice_alg} is part of a loop, which is repeated until the current $x_c$ is classified differently as $x_0$ (line 18), in which case $x_c$ is a valid counterfactual for $x_0$. In step 3 (lines 12-17), NICE identifies the best hybrid instance $x_{i,b}$ in iteration $i$, in which exactly one feature $j$ from \textit{featureList} changes its value from that in $x_c$ to that in $x_n$. Once the value of feature $j$ has been changed, the reward function is calculated (see Section \ref{sec:Reward Functions} for details), and the instance $x_{i,b}$ with the highest reward function is retained as $x_c$. Finally, NICE checks whether the new instance $x_c$ results in a class change compared to $x_0$, in which case the algorithm terminates and returns $x_c$ as a valid counterfactual. Otherwise, NICE starts a new iteration of step 3 with the updated $x_c$, after removing the currently best feature $b$ from \textit{featureList}. The beauty of this approach is that we will always end up with an explanation, since after the last iteration there is only one candidate left, which is $x_{n}$, for which we know that it is a counterfactual instance.

Finally, we want to point out that NICE offers an anytime counterfactual solution, since it can at any time return an actual counterfactual, and hence does not need to run to completion to provide a counterfactual, as required by e.g., gradient descent approaches such as DiCE.

\begin{algorithm}[h!]
	\small
	\caption{NICE (Nearest Instance Counterfactual Explanations)}\label{nice_alg}
	\textbf{NICE}($x_0$)
	\vspace{-0.5mm}
	\begin{algorithmic}[1]
	    \State \textbf{Input:} $x_0$: instance for which to find counterfactual
	    \State \textbf{Output:} $x_c$: counterfactual instance for $x_0$
		\State \textbf{Step 1: Initialization}
		\State $x_{0,b} \leftarrow$ $x_{0}$ //$x_{0,b}$ is the best instance from iteration 0
		\State $x_{c} \leftarrow$ $x_{0}$ //$x_c$ is the current hybrid instance between $x_0$ and $x_n$ 
		\State $i\leftarrow 1$ //$i$ keeps track of the current iteration of step 3
		\State \textbf{Step 2: Find nearest unlike neighbour}
		\State $x_{n} \leftarrow$ FIND-NEAREST-UNLIKE-NEIGHBOUR($x_0$)
		\State \textit{featureList} $ \leftarrow$ IDENTIFY-NON-OVERLAPPING-FEATURES($x_0,x_n$)
		\State \textbf{do}
		\State \hspace{\algorithmicindent}\textbf{Step 3: Determine best hybrid instance}
		\State \hspace{\algorithmicindent}$x_{i,b} \leftarrow \emptyset$, $R(x_{i,b})\leftarrow -\infty$, $b\leftarrow \emptyset$ //Keep track of best hybrid instance per iteration $i$
		\State \hspace{\algorithmicindent}\textbf{for} $j$ \textbf{in} \textit{featureList}:
		\State \hspace{\algorithmicindent}\hspace{\algorithmicindent}$x_j \leftarrow x_c$, $x_j[j] \leftarrow x_n[j]$
		\State \hspace{\algorithmicindent}\hspace{\algorithmicindent}\textbf{if} $R(x_j) > R(x_{i,b})$ \textbf{then} $x_{i,b} \leftarrow x_j$, $R(x_{i,b})\leftarrow R(x_j)$, $b \leftarrow j$
		\State \hspace{\algorithmicindent}$x_{c} \leftarrow x_{i,b}$ //Update current hybrid instance
		\State \hspace{\algorithmicindent}\textit{featureList} $\leftarrow$ \textit{featureList}$\setminus \{b\}$ //Remove feature $b$ 
		\State \textbf{while} $\hat{y_c}= \hat{y_0}$ //Predicted class is still the same for $x_c$ and $x_0$
		\State \textbf{return} $x_{c}$
	\end{algorithmic}
\end{algorithm}

To give the reader some idea of NICE's worst case asymptotic running time, we use $O$ (big O) notation to signify the worst case behaviour. For an in-depth introduction to asymptotic running times and algorithm complexity, we refer to Cormen et al. \cite{cormen2009}.
\begin{itemize}
    \item Lines 4-6 take $O(1)$ time since these are variable assignments.
neighbors
    \item Line 9: to identify non-overlapping features between two instances, we need to compare the values of each feature, which leads to $O(m)$ complexity.
    \item Lines 10-18 are part of a loop that is executed at worst $O(m)$ times, since in the worst case we need to consider $O(m)$ hybrid instances between $x_0$ and $x_n$. Note that since $x_n$ is a nearest unlike neighbor of $x_0$, the actual number of features with a different value is likely (considerably) smaller than $m$, e.g., $m-10$, but from an asymptotic worst case point of view, this amounts to $O(m)$ iterations of the loop.
    \begin{itemize}
        \item Line 12 takes $O(1)$ time since these are variable assignments.
        \item Lines 13-15 constitute another loop which is repeated $O(m)$ times in the worst case, since the number of features with different values is again considered just like in the outer loop.  
        \begin{itemize}
            \item Line 14 takes $O(1)$ time since these are variable assignments.
            \item Line 15 involves the calculation of $R(x_j)$ ($R(x_{i,b})$ has been calculated in a previous iteration), along with two variable assignments, which leads to $O(R(x))$ complexity.
            \item The overall time complexity of the inner loop amounts to $O(m\cdot(1+R(x)))=O(m\cdot R(x))$.
        \end{itemize}
        \item Line 16 takes $O(1)$ time since this is a variable assignment.
        \item The same applies for line 17.
        \item The overall time complexity of the outer loop amounts to $O(m\cdot(1+m\cdot R(x)+1+1))=O(m^2\cdot R(x))$.
    \end{itemize}
    \item The overall worst case asymptotic running time of NICE then constitutes $O(1+k\cdot m +m+m^2\cdot R(x))=O(R(x)\cdot m^2+k\cdot m)$.
\end{itemize}

\begin{figure*}
  \includegraphics[width=1\textwidth]{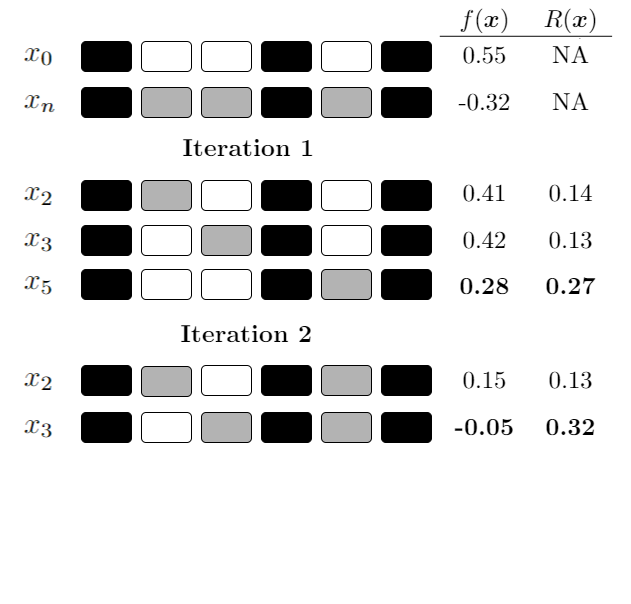}
\caption{Example of counterfactual explanations for loan approval.}
\label{NICEalgo}
\end{figure*}

An example of how NICE works on instances with six features is shown in Figure \ref{NICEalgo}. We start by selecting the nearest unlike neighbor $x_{n}$ from the training set, for which holds: $\hat{y}_0 \neq \hat{y}_{n}$ and $y_{n} = \hat{y}_{n}$. The top two rows of Figure~\ref{NICEalgo} show two data instances $x_0$ and $x_{n}$, each with six features. The black squares represent the feature values for which both instances overlap. The white and gray squares respectively represent the feature values of $x_0$ and $x_{n}$ for the remaining features. 

Next, iteration 1 shows the three hybrid instances that can be created with the first application of steps 2 and 3 from algorithm \ref{nice_alg}, in which exactly one non-overlapping feature is replaced with the corresponding value from $x_{n}$. $x_{2}$ uses the value of the second feature of $x_{n}$, $x_{3}$ uses the value of the third feature, and $x_{5}$ uses the value of the fifth feature. For each of these new hybrid instances we calculate the outcome of a reward function $R(x)$, which will be discussed in Section \ref{sec:Reward Functions}. The instance with the highest value for $R(x)$, in this case $x_5$, has the most desirable properties. We then check if this instance is predicted as the opposite class of $x_0$. If so, we have our counterfactual explanation and stop the search. In our example, this is not the case and we continue our search with $x_{5}$ as the new $x_c$.

In the next iteration, we check the non-overlapping features of the new $x_c$ and $x_{n}$. Again, we create all possible new combinations where one feature of $x_c$ is replaced by the feature value of $x_{n}$. At this point, the candidate with the highest reward function ($x_{2}$) is predicted as a different class, so now we have found a counterfactual explanation. 

\subsection{Reward Functions}\label{sec:Reward Functions}
We use three reward functions, each of which measures the effect of a perturbation on the score per unit of sparsity, proximity, or plausibility. These three reward functions amount to three different objectives for NICE and can hence be used to accommodate different user requirements. Users can then e.g., choose which of these three types of counterfactuals they prefer based on the outcomes of the three versions of NICE. In the remainder of the paper, we will call these three algorithm versions: NICE(\texttt{spars}), NICE(\texttt{prox}) and NICE(\texttt{plaus}). The worst case time complexity of each of these three versions of NICE is displayed in Table \ref{tabcompl}, and discussed in more detail below.

\subsubsection{Sparsity}
Sparsity happens to be the most straightforward property to optimize with our approach. By simply selecting the perturbation that has the highest prediction score at each iteration, we effectively optimize for sparsity, as shown in reward function (\ref{SparsLoss}).


\begin{equation}
\label{SparsLoss}
R(x) = \hat{y}\cdot\frac{f(x_{i-1,b})-f(x)}{sparsity(x_{i-1,b},x)} = \hat{y}\cdot(f(x_{i-1,b})-f(x))
\end{equation}

This function compares the score and sparsity of each candidate $x$ with that of the best candidate $x_{i-1,R_{max}}$ from the previous iteration with a factor $\hat{y}$ added to ensure a correct sign of our reward function for both classes. The sparsity difference between these instances is by definition one because we exactly add one extra feature to the explanation candidate each iteration. This allows us to remove the denominator from the formula. We then end up with a reward function which measures this score difference. This results in a theoretical range of \lbrack -2,2 \rbrack for the sparsity reward function. 
Note  that this is exactly the same as the one implicitly used by SEDC \cite{fernandez2020}. The difference is that we replace the feature values with those of $x_{n}$, while SEDC uses the mean or mode of these features. 

The time complexity of function (\ref{SparsLoss}) depends on $f(x)$, which determines the classification of instance $x$ based on the classification model $f$ used. It is crucial to consider that any classification model is trained offline and that when $f(x)$ is calculated as part of NICE, we only classify an instance $x$ according to the trained model $f$, but we do not retrain the model itself. This also means that the time complexity of the sparsity function is $O(f(x))$. The resulting worst-case time complexity for NICE as a whole is shown in Table \ref{tabcompl}.

\begin{table}
	\small
	\begin{center}
		\begin{tabular}{ll}
			\textbf{Reward function} &\textbf{Worst case time complexity NICE} \\
			\hline
			Sparsity & $O(f(x)\cdot m^2 +k\cdot m)$\\
			Proximity & $O(m^3+f(x)\cdot m^2 +k\cdot m)$\\
			Plausibility & $O((f(x)+g(x))\cdot m^2 +k\cdot m)$\\
		\end{tabular}
	\end{center}
	\caption{Overview of NICE's worst case time complexity for each reward function. $k$ is the number of instances, $m$ the number of features, $f(x)$ is the classification of an instance $x$ by the employed classification model, and $g(x)$ is the use of the AE for an instance $x$. The expressions have been rewritten such that each expression can be seen as a polynomial function in terms of $m$.}
	\label{tabcompl}
\end{table} 

\subsubsection{Proximity}
Proximity refers to the distance from the original data point $x_0$ to $x_c$. In the reward function below, we have replaced the sparsity measure of function (\ref{SparsLoss}) with a proximity measure.

\begin{equation}\label{ProxLoss}
R(x) = \hat{y}\cdot\frac{f(x_{i-1,b})-f(x)}{d(x_0,x)-d(x_0,x_{i-1,b})}
\end{equation}

This function effectively calculates the decrease in prediction score per unit of distance. Sparsity and proximity often go hand in hand and (as our results show) both optimization methods often lead to the same explanation. Each iteration we move further from $x_0$ which makes the theoretical bounds of denominator $\rbrack 0,+\infty \lbrack$. Furthermore, we know from reward function \ref{SparsLoss} that the bounds of our nominator are $\lbrack-2,2\rbrack$, which results in a theoretical bound of $\rbrack -\infty,\infty \lbrack$ for the proximity reward function.

The time complexity of function (\ref{ProxLoss}) depends on $f(x)$ and on the distance function $d$. Whereas the former once again depends on the classification model used, the latter is $O(m)$ since we have to loop over all features to compute the distance between two instances. As a result the worst case time complexity of the proximity function is $O(m+f(x))$. The total worst-case time complexity for NICE with proximity, after some rearranging of terms, is shown in Table \ref{tabcompl}.

\subsubsection{\label{sectionplaus}Plausibility}
We use the AE reconstruction error as a proxy for plausibility, which makes the employed plausibility function very similar to the metrics IM1 and IM2 used for interpretability by Van Looveren \& Klaise \cite{CFproto}. An AE uses a neural network to project an instance onto a latent space and then tries to reconstruct this instance~\cite{kramer1991nonlinear}. The error represents how successful the instance is reconstructed. When we train an AE on our training data, we can use the reconstruction error of an instance to measure how similar it is to this data. A higher (lower) error represents a data point farther from (closer to) the data manifold. 
\begin{equation}\label{PlausLoss}
R(x) = \hat{y}\cdot\frac{f(x_{i-1,b})-f(x)}{(AE_{error}(x_{i-1,b})-AE_{error}(x))^{-1}}
\end{equation}

This reward function behaves differently from the two previous ones. First, unlike the sparsity and proximity difference, the AE error difference can be negative and has theoretical bounds of $\rbrack -\infty, \infty \lbrack$. This is because it is possible that the hybrid instance has an AE error that is larger than the AE error of any of the two real observations $x_0$ and $x_n$. It is even most likely that the AE error is larger for both those instances because it is not a real observation from the dataset. As a result reward function \ref{PlausLoss} has theoretical bounds of $\rbrack -\infty, \infty \lbrack$
Second, the plausibility function depends on the AE in terms of complexity. However, the AE is trained offline and as part of the plausibility function we only compute the trained AE output given an instance $x$. To avoid an in-depth discussion of Artificial Neural Networks, of which an AE is a type, we limit ourselves to stating the AE time complexity as $O(g(x))$, with $g(x)$ referring to the AE. This results in a worst-case time complexity of $O(f(x)+g(x))$ for the plausibility reward function. Table \ref{tabcompl} contains the total worst-case time complexity of NICE with plausibility.

Despite these downsides, the results show that it is still a valid optimization strategy. Also note that reward function (\ref{SparsLoss}) for sparsity is part of the plausibility reward function (\ref{PlausLoss}). Therefore we are also optimizing for sparsity and the resulting explanation will be a balance between these two properties.

\section{\label{exp}Experiments}
\subsection{\label{testdesign}Test design}

We test NICE on datasets retrieved from the Penn Machine Learning Benchmark (PMLB) \cite{olson2017pmlb}, from which we make a selection based on three criteria. First, the dataset must contain at least 500 instances. Second, the dataset must be a binary classification task and finally, one of our classifiers must get an AUC of at least 0.6 on the test set. This results in the 40 datasets which are shown in Table \ref{TableDataSummary}. Specifically, the columns of the table contain the name of the dataset, the number of instances included, and the number of features for each instance. The latter is furthermore split into two additional columns, which show the number of categorical and numerical features respectively. Consider that there is variation between the different datasets both in the number of types of features, and in the proportion of both feature types (e.g., mostly categorical, only numerical). Furthermore, for each dataset the class imbalance, i.e., the ratio of the positively and negatively classified instances, is also included.

For each dataset we start by creating a test set which contains 20\% (with a minimum of 200 instances) of the data. The rest of the dataset is used as a training set to train both a Random Forest classifier (RF) and an Artificial Neural Network (ANN). The AUC values obtained for both classifiers are shown in the final two columns of Table \ref{TableDataSummary}. The parameters of both models are trained using a five-fold cross-validation. \footnote{See Appendix \ref{AppendixParameter} for more details about the parameter tuning.} Finally, counterfactual explanations are generated with all algorithms for a random sample of 200 instances from the test set. In total 200 counterfactual explanations are generated with 10 counterfactual algorithms for 2 classifiers on 40 datasets, resulting in a sample of 160,000 explanations. For each counterfactual, we calculate diverse metrics to assess their quality. 

In our statistical analyses we use a Friedman-rank test \cite{friedman1937use,friedman1940comparison} to evaluate the statistical significance of differences between results for different algorithms. We do this by ranking all algorithms for each observation, giving a rank of 1 to the best performing explanation algorithm and a rank of 10 to the worst. If no counterfactual is available, we give this algorithm the worst rank for this observation. If there is a tie, we use the lowest rank for all tied algorithms. We then report the averages over all datasets of these ranks and submit them to the Friedman test. If this test rejects the null hypothesis of indifferent rank means, we calculate the critical difference using a Nemenyi test \cite{nemenyi1962distribution}. If the difference in average ranks between two algorithms is greater than this critical difference, we conclude that they are significantly different. In Tables \ref{ANNMainResults}, \ref{RFMainResults}, \ref{tabres3} and \ref{tabres4} (discussed in the following subsections), the best values are marked in bold, as are values not significantly different from the best ones (5\% confidence level)\footnote{All metrics are compared over the same number of observations and algorithms, causing the critical difference to always be 0.151.}. This use of ranks solves two problems. First, it makes the comparison possible over observations from different datasets and second, it allows us to work with data points for which not all algorithms are able to generate a counterfactual explanation.

Finally, because a fair comparison is a crucial part in our research, we make sure all algorithms have access to the same resources in our experiments. Therefore, all experiments are run on an Amazon EC2 C4.8xlarge instance where each counterfactual explanation is generated on a single 2.9 GHz Intel Xeon E5-2666 v3 Processor.

\begin{table}
\footnotesize
	\begin{center}
		\begin{tabular}{lccccccccc}
		\textbf{Name} & \textbf{\#Inst.}  & \textbf{\#Feat.} & \textbf{\#Cat.}& \textbf{\#Num.}& \textbf{Class}& \textbf{AUC} & \textbf{AUC}\\
		&  &  & \textbf{feat.} & \textbf{feat.} & \textbf{imbalance} & \textbf{(ANN)} & \textbf{(RF)}\\
			\hline
            adult	&48,842&	14&	5	&9&	0.761&	0.903&	0.913\\
            agaricus\_lepiota&	8154&	22	&21	&1	&0.481&	1.000&	1.000\\
            australian	&690&	14&	7&	7&	0.445&	0.905&	0.940\\
            breast\_w&	699	&9&	8&	1&	0.345&	0.991&	0.997\\
            buggyCrx&	690&	15&	8&	7&	0.555&	0.921&	0.949\\
            chess&	3196&	36&	36&	0&	0.522&	1.000&	0.999\\
            churn&	5000&	20&	4&	16&	0.142&	0.872&	0.919\\
            clean2&	6598&	168	&0	&168&	0.154&	1.000&	1.000\\
            coil2000&	9822&	85&	84&	1&	0.060&	0.691&	0.745\\
            credit\_a&	690&	15&	8&	7&	0.555&	0.902&	0.910\\
            credit\_g&	1000&	20&	17&	3&	0.700&	0.663&	0.731\\
            crx&	690&	15&	8&	7&	0.445&	0.859&	0.941\\
            diabetes&	768&	8&	0&	8&	0.349&	0.823&	0.851\\
            dis&	3772&	29&	23&	6&	0.985&	0.895&	0.989\\
			GAMETES1&	1600&	20&	20&	0&	0.500&	0.636&	0.648\\
			GAMETES2&	1600&	20&	20&	0&	0.500&	0.746&	0.780\\
            GAMETES3&	1600&	20&	20&	0&	0.500&	0.664&	0.722\\
            GAMETES4&	1600&	20&	20&	0&	0.500&	0.690&	0.705\\
            german&	1000&	20&	17&	3&	0.700&	0.718&	0.758\\
            Hill\_Valley	&1212	&100	&0&	100&	0.505&	0.993&	0.557\\
            hypothyroid&	3163&	25&	18&	7&	0.952&	0.975&	0.988\\
            kr\_vs\_kp&	3196&	36&	36&	0&	0.522&	1.000&	0.999\\
            magic&	19,020&	10&	0&	10&	0.352&	0.922&	0.937\\
            mofn\_3\_7\_10&	1324&	10&	10&	0&	0.779&	1.000&	1.000\\
            monk1&	556&	6&	6&	0&	0.500&	1.000&	1.000\\
            monk2&	601&	6&	6&	0&	0.342&	1.000&	0.896\\
            monk3&	554&	6&	6&	0&	0.520&	0.992&	0.986\\
            mushroom&	8124&	22&	21&	1&	0.482&	1.000&	1.000\\
            parity5+5&	1124&	10&	10&	0&	0.504&	1.000&	0.674\\
            phoneme&	5404&	5&	0&	5&	0.294&	0.906&	0.970\\
            pima&	768&	8&	0&	8&	0.349&	0.867&	0.819\\
            profb&	672&	9&	3&	6&	0.333&	0.633&	0.676\\
            ring&	7400&	20&	0&	20&	0.505&	0.990&	0.992\\
            spambase&	4601&	57&	0&	57&	0.394&	0.974&	0.988\\
            threeOf9&	512&	9&	9&	0&	0.465&	0.972&	0.999\\
            tic\_tac\_toe&	958&	9&	9&	0&	0.653&	0.997&	1.000\\
            tokyo1&	959&	44&	2&	42&	0.639&	0.962&	0.983\\
            twonorm&	7400&	20&	0	&20	&0.500&	0.996&	0.997\\
            wdbc&	569&	30&	0&	30&	0.371&	0.973&	0.981\\
            xd6&	973&	9&	9&	0&	0.331&	1.000&	1.000\\
		\end{tabular}
	\end{center}
\caption{Descriptive statistics and performance metrics of all binary datasets.}
\label{TableDataSummary}
\end{table}

\subsection{Algorithmic requirements}
First, we check if all reviewed algorithms meet the algorithmic requirements of Sokol \& Flach \cite{sokol2020explainability}. These properties are often overlooked, but are actually very important, as they determine whether algorithms can be implemented in real-world applications. The three main properties are access, coverage and computation time \cite{sokol2020explainability}. We go over each of them in the following paragraphs based on the results from Tables \ref{ANNMainResults} and \ref{RFMainResults}\footnote{Whereas these and other tables contain a summary of our results across the different datasets of Table \ref{TableDataSummary}, a detailed overview of all results per dataset can be found in the online appendix (https://github.com/DBrughmans/NICE\_experiments).}.

The required \textbf{access} of all these algorithms is the same. Each needs access to the training data and the scoring output of the classification model. 

\textbf{Coverage} refers to the percentage of instances for which a valid counterfactual explanation is found. The reported coverages in Tables \ref{ANNMainResults} and \ref{RFMainResults} show the strength of NICE's design. By construction, all versions of NICE have 100\% coverage. In reality, anything less than perfect coverage is often not accepted. However, of the other algorithms, only WIT, which is very similar to NICE(\texttt{none}), has a perfect coverage. For both the ANN and RF, GeCo is the worst algorithm with an average coverage of only 32.4\% and 35.5\% respectively, while the other benchmark algorithms have a coverage between 62.7\% (CBR) and 78.4\% (SEDC) for the ANN, and between 48.3\% (CBR) and 72.0\% (SEDC) for the RF. Consider that SEDC always has perfect coverage on one of both classes (recall that we are working with binary datasets). SEDC replaces feature values with their mean or mode, and after the maximum number of replacements it ends with an instance consisting of only means or modes. If we are looking for a counterfactual instance that belongs to the predicted class of this instance, it will therefore always be found. In some applications such as fraud detection, credit scoring and clinical healthcare, we are mostly interested in explanations for one class. If this matches the class with perfect coverage for SEDC, it is a valid option. The other algorithms with imperfect coverage are unpredictable in the instances for which no explanation can be created, which makes them unusable in many real-world applications where the algorithm should be able to generate an explanation for all predictions.

\begin{table}[h]
	\footnotesize
	\begin{center}
		\begin{tabular}{l|cccc|cccccc}
			&\textbf{NICE} & \textbf{NICE} & \textbf{NICE} & \textbf{NICE} & \textbf{WIT} & \textbf{CBR} & \textbf{SEDC} & \textbf{DiCE} & \textbf{CFproto} & \textbf{GeCo}\\
			&(\texttt{none}) & (\texttt{spars}) & (\texttt{prox}) & (\texttt{plaus})\\
			\hline
			Coverage (\%) & 100.0& 100.0& 100.0& 100.0& 100.0& 62.7& 78.4& 70.1& 76.7& 49.4\\
			CPU time (ms)& 3.7& 14.8& 20& 81.3& 6.1& 26.2& 45.7& 130,783.0& 19,315.8& 1606.5\\
			\hline
			Time rank& \bf{1.26}& 4.32& 5.31& 6.6& 2.42& 4.49& 3.65& 8.27& 9.88& 8.8 \\
			Spars. rank &5.17& \bf{1.42}& 1.67& 3.33& 5.33& 4.39& 3.16& 3.67& 5.35& 7.43\\
			Prox. rank & 5.10& \bf{1.86}& \bf{1.72}& 3.48& 5.34& 5.34& 4.10& 4.85& 4.08& 7.95\\
			Plaus. rank & \bf{3.10}& 4.07& 4.15& 3.27& \bf{3.11}& 6.66& 5.31& 6.61& 5.86& 6.09\\
		\end{tabular}
	\end{center}
	\caption{Comparison of four versions of NICE and six benchmarks from the literature based on an \textbf{ANN classifier} (best values marked in bold).}
	\label{ANNMainResults}
\end{table} 

\begin{table}[h]
	\footnotesize
	\begin{center}
		\begin{tabular}{l|cccc|cccccc}
			&\textbf{NICE} & \textbf{NICE} & \textbf{NICE} & \textbf{NICE} & \textbf{WIT} & \textbf{CBR} & \textbf{SEDC} & \textbf{DiCE} & \textbf{CFproto} & \textbf{GeCo}\\
			&(\texttt{none}) & (\texttt{spars}) & (\texttt{prox}) & (\texttt{plaus})\\
			\hline
			Coverage (\%)& 100.0& 100.0& 100.0& 100.0& 100.0& 48.3& 72.0& 63.3& 68.2& 39.4\\
			CPU time (ms) & 61.8& 573.4& 925.8& 2090.6& 95.2& 320.6& 1499.6& 2,016,936.8& 799,339.6& 6478.1\\
			\hline
			Time rank & \bf{1.01}& 4.50& 5.79& 6.80& 2.04& 4.09& 4.50& 7.81& 9.71& 8.73 \\
			Spars. rank & 4.73& \bf{1.31}& 1.87& 3.15& 4.87& 5.10& 3.34& 4.11& 5.51& 7.65\\
			Prox. rank& 4.56& \bf{1.63}& \bf{1.63}& 3.16& 4.88& 5.85& 4.24& 5.50& 4.52& 7.89\\
			Plaus. rank& \bf{2.91}& 4.10& 4.13& \bf{2.88}& \bf{2.93}& 6.67& 5.16& 6.43& 5.97& 6.50 \\
		\end{tabular}
	\end{center}
	\caption{Comparison of four versions of NICE and six benchmarks from the literature based on an \textbf{RF classifier} (best values marked in bold).}
	\label{RFMainResults}
\end{table} 

Tables \ref{ANNMainResults} and \ref{RFMainResults} also show the \textbf{computation time} required to generate a counterfactual explanation. We show both the average time in milliseconds (CPU time) and the average rank (Time rank). First, we notice that there is a big difference in CPU times between explanations generated for an ANN and an RF. As seen in Table \ref{TableDataSummary}, making a prediction with an RF is much more computationally expensive than with an ANN. When comparing the different algorithms we notice that WIT and NICE(\texttt{none}), the two algorithms that use a nearest unlike neighbour without further optimization as a counterfactual explanation, are the fastest. The three versions of NICE with further optimization have reasonable optimization times which are all below 2.09 seconds on average. As expected, NICE(\texttt{spars}) is the fastest followed by NICE(\texttt{prox}) and NICE(\texttt{plaus}). This is due to NICE(\texttt{prox}) (NICE(\texttt{plaus})) needing to compute the distance (AE error) at each iteration. Of the other benchmark algorithms, CBR, SEDC and GeCo also have reasonable CPU times. On the contrary, DiCE and CFproto take on average several minutes to generate an explanation\footnote{The speed of CFproto and DiCE could be improved if access was given to the gradients of the ANN and RF \cite{CFproto}. But to level the playing field we used the model-agnostic version of both these algorithms in all our experiments, since all other algorithms are model-agnostic.}. As a result, we conclude that except for CFproto and DiCE, all of these algorithms are fast enough to be used in real-time applications.

With respect to NICE, recall that the worst case time complexity (Table \ref{tabcompl} in Section \ref{sec:Reward Functions}) is related to both the $k$ and $m$ values of the treated dataset. To demonstrate the impact of different values on CPU time and argue that even for larger values this does not lead to excessive CPU times, we analyze the time complexity of two datasets from Table \ref{TableDataSummary} as examples in Appendix \ref{appendixcompl}.

Based on these findings, we can conclude that all versions of NICE and WIT have the most desirable algorithmic properties. Perfect coverage ensures that these algorithms can be used in applications where explanations are required by law, such as under the Fair Credit Reporting Act~\cite{FCRA}. All versions of NICE also have an efficient run time. This is a must for high-stakes decision making under time pressure like fraud detection, credit scoring and clinical healthcare. The detailed results\footnote{See online appendix on GitHub: https://github.com/DBrughmans/NICE\_experiments} show that all versions of NICE scale well with the number of features. Even for the largest dataset with over 162 features, the slowest version (NICE(\texttt{plaus})) has an average run time of 17.4 seconds and a maximum of 23.5 seconds. This makes NICE useful in domains where predictions are made at high frequency and scalability is a priority. Taking into account all algorithmic requirements, we conclude that NICE and WIT are the best options for any generic classification model. The other algorithms are useful in specific situations. 

\subsection{Explanation requirements}
\subsubsection{\label{resbasic}Overview results}

We compare the four versions of NICE with the six selected benchmark algorithms on three counterfactual properties: sparsity, proximity and plausibility (shown as ``Spars.\ rank'', ``Prox.\ rank'' and ``Plaus.\ rank'' respectively in both Tables \ref{ANNMainResults} and \ref{RFMainResults}).

First, we look at \textbf{sparsity}. The main aim of SEDC, CBR and NICE(\texttt{spars}) is to generate the most sparse counterfactual possible. The latter clearly appears to be the winner here, since for both classification models it has by far the lowest average rank (1.36 for ANN and 1.30 for RF). GeCo, CFproto and CBR are the algorithms that perform the worst, while DiCE, SEDC, WIT and NICE(\texttt{plaus}) have an intermediate performance. Consider that NICE(\texttt{prox}) also obtains excellent sparsity results, something which we come back to in Section \ref{sectrade} and which we already hinted at in Section \ref{sectionplaus}.

Next, we consider \textbf{proximity}. All algorithms have some sort of proximity constraint, be it in direct form or induced by a sparsity constraint. The best performing algorithms in terms of proximity are NICE(\texttt{prox}) and NICE(\texttt{spars}) for both classification models. NICE(\texttt{prox}) performs slightly better but the difference is not significant. However, both algorithms are significantly better in terms of proximity compared to all other algorithms. NICE(\texttt{plaus}) also performs significantly better than the six benchmark algorithms. NICE(\texttt{none}) does considerably worse, and is also outperformed by SEDC, DiCE and CFproto, which shows how challenging it is to find an explanation with low proximity in the instances of the training set. Related to the benchmark algorithms, it is worth noting that CFproto performs considerably better for proximity than it did for sparsity, whereas the opposite holds for DiCE.

Finally, we examine the \textbf{plausibility} of all counterfactual explanations. Recall that both NICE(\texttt{plaus})'s and CFproto's main goal is to generate explanations which lie close to the data manifold. For an ANN, NICE(\texttt{none}) and WIT are better than the other algorithms, but there is no significant difference between them, whereas for an RF both are additionally tied with NICE(\texttt{plaus}). The comparatively good plausibility results for both NICE(\texttt{none}) and WIT are due to both techniques resulting in an actual instance, a nearest unlike neighbour, whereas for the other algorithms there is no such guarantee. Furthermore, NICE(\texttt{none}) only looks for counterfactual instances among correctly classified observations of the training set, which avoids areas of uncertainty in the feature space. As we have seen before, this comes with a cost of proximity and a benefit in computing time, while also allowing for a good plausibility score. 

\subsubsection{Impact performance metrics}

We previously compared all algorithms on three main requirements for the explanations: sparsity, proximity and plausibility. Whereas sparsity is well defined, there are alternatives for proximity and plausibility.

For proximity we compare four different distance metrics. The L1 distance with standardization for each feature (L1(stand.)) which is used in NICE, the L1 distance with normalization (L1 (norm)) used in WIT, the L2 distance with standardization (L2(stand.)) used in DiCE and GeCo and the L2 distance with normalization (L2(norm.)). 

Plausibility is less straightforward to measure. The AE error is a good proxy but will bias the comparison in favor of NICE(\texttt{plaus}), as it is the only algorithm that optimizes for this metric. Therefore we have added four more metrics to measure plausibility:  the average distance to the 5 nearest neighbors (5NN), IM1 \cite{CFproto}, IM2 \cite{CFproto} and a measure taken from \cite{pawelczyk2020counterfactual} which we call cross-model robustness. It is argued that if counterfactual instances respect the data manifold, they are less vulnerable to classification model uncertainty or changes over time~\cite{pawelczyk2020counterfactual}. We can measure this by checking the percentage of instances for which an explanation is also valid for another classification model trained on the same data. We use two classification models in our experiments, so we can easily check whether an explanation for one model is also an explanation for the other.

The results for all 10 algorithms given these alternative proximity and plausibility metrics are shown in Tables \ref{tabres3} and \ref{tabres4}. The metrics previously used in Section \ref{resbasic} are marked with an asterisk (*). In general, the results of both tables are in line with those of Section \ref{resbasic}: NICE(\texttt{spars}) and NICE(\texttt{prox}) have the best results irrespective of the proximity metric used, while NICE(\texttt{none}) and WIT have the best results for plausibility, though NICE(\texttt{plaus}) is again a close third, especially for the RF classifier. Notice that a small difference occurs between the two L1 distances for NICE(\texttt{none}) and WIT where each scores significantly better in their own proximity metric. For plausibility the difference between NICE(\texttt{none}) and NICE(\texttt{plaus}) becomes more pronounced, since for all alternative plausibility measurements, NICE(\texttt{none}) is significantly better than NICE(\texttt{plaus}). This again emphasizes that using actual instances from the dataset (i.e., nearest unlike neighbors) is a valid solution when only plausibility is preferred.

\begin{table}
	\footnotesize
	\begin{center}
		\begin{tabular}{ll|cccc|cccccc}
			& &\textbf{NICE} & \textbf{NICE} & \textbf{NICE} & \textbf{NICE} & \textbf{WIT} & \textbf{CBR} & \textbf{SEDC} & \textbf{DiCE} & \textbf{CFproto} & \textbf{GeCo}\\
			& &(\texttt{none}) & (\texttt{spars}) & (\texttt{prox}) & (\texttt{plaus})\\
			\hline
			\textbf{Prox.} 
			    &L1(stand.) rank *& 5.37& \bf{1.87}& \bf{1.80}& 3.63& 4.96& 5.19& 3.88& 5.21& 4.00& 7.90\\
                &L1(norm.) rank& 5.10& \bf{1.86}& \bf{1.72}& 3.48& 5.34& 5.34& 4.10& 4.85& 4.08& 7.95\\
                &L2(stand.)rank& 5.10& \bf{1.96}& \bf{1.88}& 3.48& 4.63& 5.57& 4.01& 5.78& 3.71& 7.70\\
                &L2(norm.)rank& 4.83& \bf{1.89}& \bf{1.76}& 3.3& 5.09& 5.7& 4.28& 5.34& 3.83& 7.79\\
			\hline
			\textbf{Plaus.} 
                &AE loss rank*& \bf{3.10}& 4.07& 4.15& 3.27& \bf{3.11}& 6.66& 5.31& 6.61& 5.86& 6.09\\
                &IM1 rank& \bf{3.49}& 4.31& 4.38& 3.89& \bf{3.57}& 6.06& 5.69& 4.62& 6.49& 5.76\\
                &IM2 rank& \bf{2.46}& 5.28& 5.41& 3.75& \bf{2.54}& 6.40& 6.00& 4.55& 6.56& 5.39\\
                &5NN rank& \bf{1.92}& 3.86& 3.92& 2.76& \bf{1.96}& 6.45& 5.82& 6.92& 6.02& 5.76\\
                &CM robustness (\%)& \bf{94.3}& 68.3& 68.7& 81.3& 85.6& 43.4& 45.6& 54.9& 43.5& 47.6\\

		\end{tabular}
	\end{center}
	\caption{Average ranks of four versions of NICE and six benchmarks from the literature for different proximity and plausibility metrics based on an \textbf{ANN classifier}. CM robustness stands for Cross-Model robustness. Standardized (stand.) and normalized (norm.) refers to how the numerical features are scaled before calculating the respective distance metric. For all ranked metrics the null-hypothesis of indifferent rank means is rejected at a 5\% significance level and the critical difference equals 0.151. Best ranks and those that are not significant different from it are marked in bold. For CM robustness, the best value is marked in bold.}
	\label{tabres3}
\end{table} 

\begin{table}
	\footnotesize
	\begin{center}
		\begin{tabular}{ll|cccc|cccccc}
			& &\textbf{NICE} & \textbf{NICE} & \textbf{NICE} & \textbf{NICE} & \textbf{WIT} & \textbf{CBR} & \textbf{SEDC} & \textbf{DiCE} & \textbf{CFproto} & \textbf{GeCo}\\
			& &(\texttt{none}) & (\texttt{spars}) & (\texttt{prox}) & (\texttt{plaus})\\
			\hline
			\textbf{Prox.} 
			    &L1(stand.) rank*& 4.85& \bf{1.69}& \bf{1.76}& 3.35& 4.48& 5.75& 4.08& 5.57& 4.47& 7.85\\
                &L1(norm.) rank& 4.56& \bf{1.63}& \bf{1.63}& 3.16& 4.88& 5.85& 4.24& 5.50& 4.52& 7.89\\
                &L2(stand.) rank& 4.65& \bf{1.81}& \bf{1.82}& 3.27& 4.25& 5.91& 4.24& 5.91& 4.28& 7.71\\
                &L2(norm.) rank& 4.39& \bf{1.70}& \bf{1.66}& 3.06& 4.69& 6.01& 4.39& 5.80& 4.36& 7.80\\
			\hline
			\textbf{Plaus.} 
			    &AE loss rank*& \bf{2.91}& 4.10& 4.13& \bf{2.88}& \bf{2.93}& 6.67& 5.16& 6.43& 5.97& 6.50\\
                &IM1 rank& \bf{3.49}& 4.15& 4.21& 3.94& \bf{3.51}& 6.33& 5.44& 4.14& 6.18& 6.33\\
                &IM2 rank& \bf{2.39}& 5.17& 5.20& 3.55& \bf{2.43}& 6.55& 5.84& 4.31& 6.38& 5.94\\
                &5NN rank& \bf{1.77}& 3.90& 3.88& 2.54& \bf{1.81}& 6.49& 5.60& 6.76& 6.10& 6.22\\
                &CM robustness (\%)& \bf{80.9}& 64.4& 62.8& 69.6& 78.7& 40.4&50.3 & 76.0& 44.7& 42.2\\
		\end{tabular}
	\end{center}
	\caption{Average ranks of four versions of NICE and six benchmarks from the literature for different proximity and plausibility metrics based on an \textbf{RF classifier}. CM robustness stands for Cross-Model robustness. Standardized (stand.) and normalized (norm.) refers to how the numerical features are scaled before calculating the respective distance metric. For all ranked metrics the null-hypothesis of indifferent rank means is rejected at a 5\% significance level and the critical difference equals 0.151. Best ranks and those that are not significant different from it are marked in bold. For CM robustness, the best value is marked in bold.}
	\label{tabres4}
\end{table} 

To gain more insight into how similar these different metrics are, we also calculate the correlations of the ranks over all algorithms and for both classifiers. Table \ref{tabcorrelation1} contains the correlation values for sparsity and the different proximity metrics, whereas Table \ref{tabcorrelation2} holds the correlation values for the different plausibility metrics. Once again, the proximity and plausibility metrics that NICE optimizes have been marked with an asterisk.

From Table \ref{tabcorrelation1} we conclude that the different proximity metrics are highly correlated (all values $>0.900$), which implies that our choice for a specific proximity metric (L1(stand.)) did not have a large impact. Furthermore, also sparsity appears to be (highly) correlated with the proximity metrics, which means that optimizing for either sparsity or proximity is beneficial for both. However, note that the correlations with sparsity are somewhat lower than those between the different proximity metrics.

In Table \ref{tabcorrelation2} we see a positive correlation between the different plausibility metrics as well, though it is smaller than between the proximity metrics. From this we conclude that for plausibility it may still be worth to consider multiple metrics, to at least have an idea of the differences in performance, even though there is a medium positive correlation between the metrics.

\begin{table}[h]
	\footnotesize
	\begin{center}
		\begin{tabular}{l|ccccc}
			& Spars.& L1(norm.) & L1(stand.)* & L2(norm.) & L2(stand.)\\
			\hline
			Spars. & &0.851	&0.850	&0.819	&0.807\\
	    	L1(norm.) & 0.832&	&0.969&	0.975&	0.943\\
			L1(stand.)* & 0.826&	0.953& &0.946&	0.968\\
			L2(norm.) & 0.782&	0.955&	0.909& &0.955	\\
			L2(stand.) & 0.764& 0.904&	0.949&	0.929& \\
		\end{tabular}
	\end{center}
	\caption{Correlations of ranks between sparsity and different proximity metrics over all algorithms. Values below (above) the main diagonal are for the ANN (RF).}
	\label{tabcorrelation1}
\end{table} 

\begin{table}[h]
	\footnotesize
	\begin{center}
		\begin{tabular}{l|cccc}
			& AE loss*&IM1 &IM2 & 5NN\\
			\hline
			AE loss* & &0.510	&0.587	&0.748\\
			IM1 & 0.452	& &0.596 &0.523\\
			IM2 & 0.527	&0.555&	&0.660\\
			5NN & 0.715	&0.497	&0.600	& \\
		\end{tabular}
	\end{center}
	\caption{Correlations of ranks between different plausibility metrics over all algorithms. Values below (above) the main diagonal are for the ANN (RF).}
	\label{tabcorrelation2}
\end{table} 

\subsubsection{\label{sectrade}Trade-offs between counterfactual properties}

\begin{figure}
    \centering
    \begin{subfigure}[t]{0.49\textwidth}
        \centering
        \includegraphics[width=\textwidth]{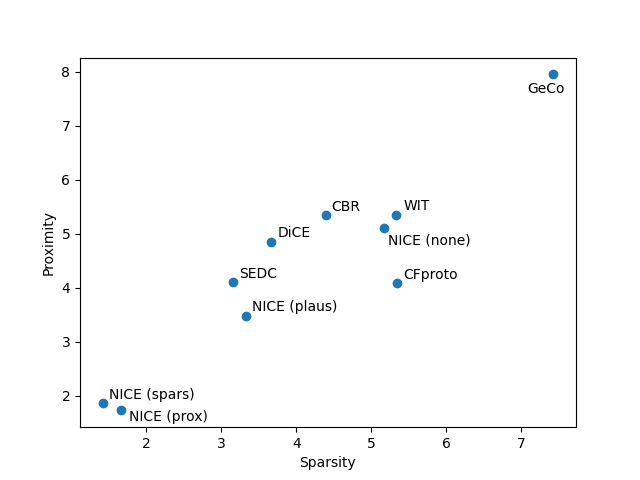}
        \caption{ANN classifier.}
        \label{FigSparsProxANN}
    \end{subfigure}
    \hfill
    \begin{subfigure}[t]{0.49\textwidth}
        \centering
        \includegraphics[width=\textwidth]{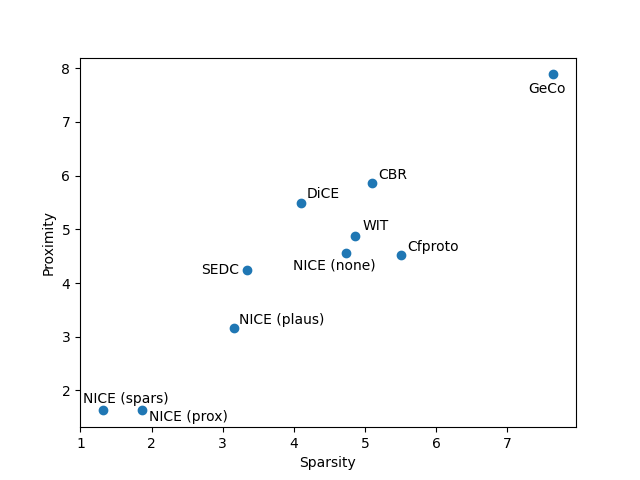}
        \caption{RF classifier.}
        \label{FigSparsProxRF}
    \end{subfigure}
    \caption{Comparison of algorithms for both proximity and sparsity average rankings (lower is better).}
    \label{Figres1} 
\end{figure}

\begin{figure}
    \centering
    \begin{subfigure}[t]{0.49\textwidth}
        \centering
        \includegraphics[width=\textwidth]{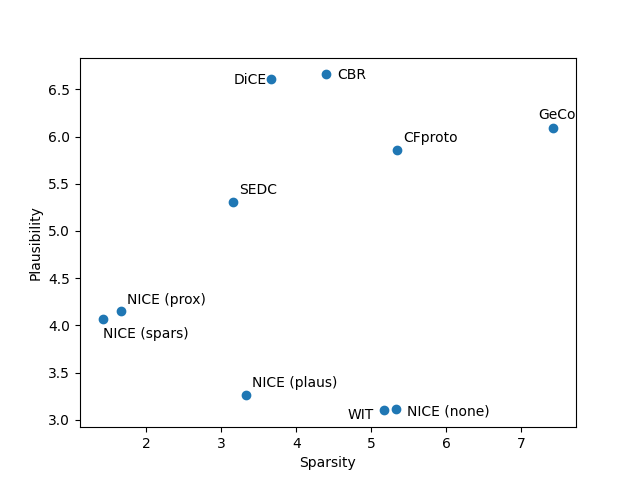}
        \caption{ANN classifier.}
        \label{FigSparsPlausANN}
    \end{subfigure}
    \hfill
    \begin{subfigure}[t]{0.49\textwidth}
        \centering
        \includegraphics[width=\textwidth]{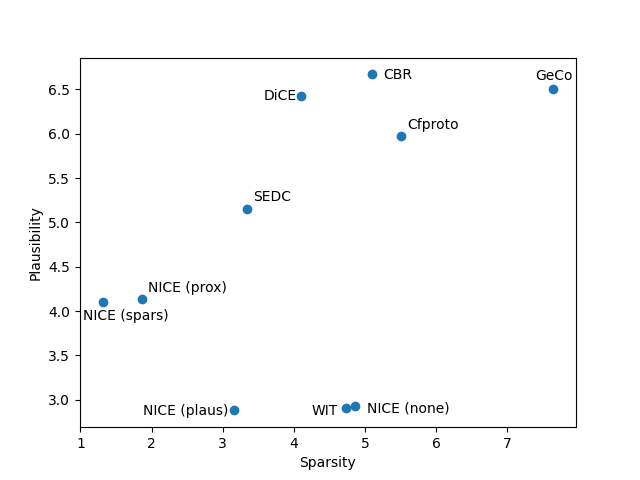}
        \caption{RF classifier.}
        \label{FigSparsPlausRF}
    \end{subfigure}
    \caption{Comparison of algorithms for both sparsity and plausibility average rankings (lower is better).}
    \label{Figres2} 
\end{figure}

\begin{figure}
    \centering
    \begin{subfigure}[t]{0.49\textwidth}
        \centering
        \includegraphics[width=\textwidth]{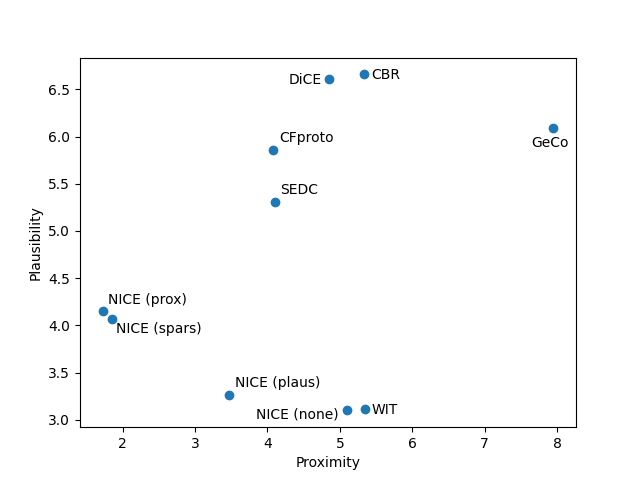}
        \caption{ANN classifier.}
        \label{FigproxPlausANN}
    \end{subfigure}
    \hfill
    \begin{subfigure}[t]{0.49\textwidth}
        \centering
        \includegraphics[width=\textwidth]{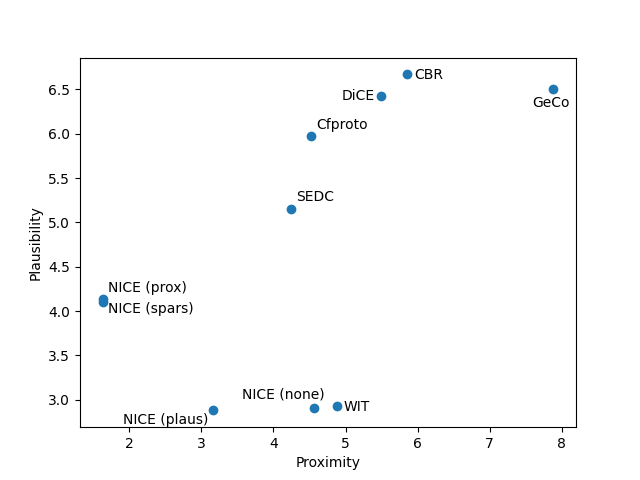}
        \caption{RF classifier.}
        \label{FigProxPlausRF}
    \end{subfigure}
    \caption{Comparison of algorithms for both proximity and plausibility average rankings (lower is better).}
    \label{Figres3} 
\end{figure}

In Figures \ref{Figres1}-\ref{Figres3} we compare all variants of NICE with the benchmarks by showing the average ranks of each algorithm for each pairwise combination of the counterfactual properties, and this for both classifiers. In this way, we determine whether some algorithms are dominated by others for a specific pair of counterfactual properties or not. Consider that an algorithm is dominated if its average rank is higher than or equal to the average rank of another for the two properties displayed. 

\begin{itemize}
    \item Sparsity-proximity (Figure \ref{Figres1}): NICE(\texttt{spars}) and NICE(\texttt{prox}) dominate all others, as they have a lower average rank for sparsity and proximity than all other algorithms. Graphically this means all others are situated ``to the right of'' and ``higher than'' NICE(\texttt{spars}) and NICE(\texttt{prox}). Of the latter two, neither dominates the other since NICE(\texttt{spars}) does better for sparsity and NICE(\texttt{prox}) better for proximity, though the difference in proximity rank is very small, especially for the RF classifier, and is furthermore not significant at a 5\% confidence level (see Tables \ref{ANNMainResults} \& \ref{RFMainResults}).
    \item Sparsity-plausibility (Figure \ref{Figres2}): NICE(\texttt{spars}), NICE(\texttt{plaus}) and WIT are not dominated by any other, hence, out of the 10 algorithms under study these three constitute an efficient frontier, in which case any of these are valid options depending on a user's preference for either better sparsity or plausibility. All other algorithms are dominated by at least one of these three.
    \item Proximity-plausibility (Figure \ref{Figres3}): the four variants of NICE dominate all other algorithms, although we should distinguish between, on the one hand NICE(\texttt{spars}) and NICE(\texttt{prox}), which have very good results for proximity but less good for plausibility, and on the other hand NICE(\texttt{plaus}) and NICE(\texttt{none}), for which the opposite holds. Although the differences between NICE(\texttt{spars}) and NICE(\texttt{prox}) are again small, the four versions of NICE constitute an efficient frontier.
\end{itemize}

In general, we can conclude that Figures \ref{Figres1}-\ref{Figres3} demonstrate the edge NICE has compared to existing approaches from the literature, though which version of NICE performs best depends on the combination of counterfactual properties. We notice that between the versions of NICE there is a clear trade-off between plausibility on the one hand and proximity or sparsity on the other hand. NICE(\texttt{none}) generates very plausible explanations at the cost of proximity and sparsity, while this is the other way around for NICE(\texttt{prox}) and NICE(\texttt{spars}). NICE(\texttt{plaus}) seems to offer a middle ground.

In Table \ref{adultexample} we show an example of how the counterfactual explanations differ for each version of NICE. The example uses the adult dataset, where we want to predict if a person's income is above or below \$ 50,000 a year based on demographical properties. In our example, the instance to explain ($x_0$) is classified as having an income below this threshold. Both NICE(\texttt{spars}) and NICE(\texttt{prox}) provide the same counterfactual explanation with a sparsity of 1 by suggesting to only increase capital-gains with \$ 5178. NICE(\texttt{plaus}) suggest an additional increase in age with two years. It is very likely that people with higher capital-gains are also older in the adult dataset, and therefore NICE(\texttt{plaus}) suggests an additional change that brings the counterfactual instance closer to the data-manifold. Finally, there is NICE(\texttt{none}, which represents a real instance from the dataset and suggests an additional reduction of the number of working-hours per week of 10.

\begin{table}[H]
	\small
	\begin{center}
    \begin{tabular}{l|l|llll}
                            & \textbf{$x_0$} & \textbf{NICE(\texttt{none})} & \textbf{NICE(\texttt{spars})} & \textbf{NICE(\texttt{prox})} & \textbf{NICE(\texttt{plaus})} \\ \hline
    \textbf{capital-gain (\$)}   & 0              & \textbf{5178}       & \textbf{5178}        & \textbf{5178}       & \textbf{5178}        \\
    \textbf{age}            & 52             & \textbf{54}         & 52          & 52                  & \textbf{54}          \\
    \textbf{hours-per-week} & 60             & \textbf{50}         & 60                   & 60                  & 60                   \\
    \textbf{workclass}      & Private        & Private             & Private              & Private             & Private              \\
    \textbf{marital-status} & Divorced       & Divorced            & Divorced             & Divorced            & Divorced             \\
    \textbf{relationship}   & Unmarried      & Unmarried           & Unmarried            & Unmarried           & Unmarried            \\
    \textbf{race}           & White          & White               & White                & White               & White                \\
    \textbf{sex}            & Female         & Female              & Female               & Female              & Female               \\
    \textbf{education}      & Bachelors        & Bachelors              & Bachelors              & Bachelors             & Bachelors              \\
    \textbf{education-num}  & 9              & 9                   & 9                    & 9                   & 9                    \\
    \textbf{occupation}     & Sales          & Sales               & Sales                & Sales               & Sales                \\
    \textbf{capital-loss (\$)}   & 0              & 0                   & 0                    & 0                   & 0                    \\
    \textbf{native-country} & US             & US                  & US                   & US                  & US                  
    \end{tabular}
    \end{center}
    \caption{Example of a counterfactual instances for each version of NICE for one instance $x_0$ of the adult dataset. The suggested changes for each counterfactual instance is marked in bold.}
	\label{adultexample}
\end{table}

\section{\label{concl}Conclusions \& future research avenues}
\subsection{Conclusions}
In this paper, we have introduced NICE, a new state-of-the-art algorithm for generating counterfactual explanations for heterogeneous tabular data. NICE is able to simultaneously achieve 100\% coverage, model-agnosticism and fast counterfactual generation for different types of classification models, thereby making it suitable for real-world applications, where these algorithmic properties are expected. Specifically, NICE starts from a nearest unlike neighbour, an existing instance correctly classified as belonging to the opposite class and subsequently includes feature values from this instance in the instance to be explained, one feature at a time, until a class change occurs.

In the extensive computational experiments, we have shown that NICE outperforms existing counterfactual algorithms from the literature for sparsity, proximity and plausibility objectives. Furthermore, when we consider alternative proximity and plausibility metrics from the literature, to determine how (un)similar these are, NICE still obtains the best results. We also looked into trade-offs between the counterfactual properties (sparsity, proximity and plausibility) and conclude that the version of NICE which performs best depends on the combination of counterfactual properties under consideration. That being said, our results show that a strong correlation exists between sparsity and proximity, which means that the most important trade-off occurs between sparsity and proximity on the one hand and plausibility on the other hand. Based on the explanations provided, users can then choose which of these three types of counterfactuals they prefer. 

\subsection{Future research avenues}
Though the goal of our paper is to propose a new algorithm, we should not ignore the importance of user expectations. Although some studies argue in favor of sparse explanations, others conclude that user expertise is paramount in determining what explanations should look like (Section \ref{user}). More specifically, future research should then test algorithms such as NICE in real-world settings, and use feedback from all stakeholders in the predictive decision making process. A multidisciplinary approach with data scientists, domain experts and end users is needed to further improve the properties of counterfactual explanations, and see for which applications they are most valuable. The results of such user studies can subsequently help to decide which version of NICE (or other algorithm, or even which type of explanation aside not limited to only counterfactuals) would be the most suited for which application.

Another future research avenue concerns the data types used, since it would be interesting to test if NICE’s approach could be extended to different data types such as behavioral, textual or image data. However, different data types bring about different levels of interpretability for a person \cite{guidotti2018}, different data properties, different types of AI techniques to be used, and due to this, also different treatment in the generation and evaluation of counterfactual methods. For example, image data is typically highly dimensional, which implies that these cannot be iterated over in a reasonable time (step 3 of Algorithm \ref{nice_alg}). One way around this could be to group pixels from images into meaningful sections \cite{vermeire2022explainable}. For textual and behavioral data, adding features is often a more radical change than removing features. To make NICE compatible with this data, perturbations should perhaps only be limited to features which are already present in the instance to explain.

Regarding the type of datasets, we have restricted ourselves to binary ones, in which the class prediction only has two options, i.e., belonging to the class we are interested in or not. However, our approach can be generalized to multiclass classification with a small change to the reward functions, which we discuss in Appendix \ref{Appendixmulticlass}. This approach would also result in a perfect coverage, however further experiments have to be done to check if these explanations also have desirable counterfactual properties.

Finally, in our experiments we only compare with other counterfactual algorithms, though one could argue that a broader comparison with other XAI methods is warranted. However, as shown by Fern\'{a}ndez-Lor\'{i}a et al. \cite{fernandez2020} in their comparison of counterfactuals with feature importance explanations, features with a large impact on \textit{prediction} may not necessarily be relevant for the \textit{explanation} of specific decisions. Therefore, the results of feature importance methods may be misleading. Despite the above, it would be worthwhile to compare counterfactual algorithms with other XAI methods, including feature importance, if only to become aware of pitfalls.

\clearpage
\printbibliography
\clearpage  

\newpage
\appendix
\section{Appendix}

\subsection{\label{appendixcompl}Examples time complexity}

With respect to NICE, recall that the worst case time complexity (Table \ref{tabcompl} in Section \ref{sec:Reward Functions}) was related to both the $k$ and $m$ values of the treated dataset. More specifically, let us consider two datasets from Table \ref{TableDataSummary}, namely ``adult'' where $k=0.8*48,842=39,073$ and $m=14$ and ``clean2'' where $k=6598*0.8=5278$ and $m=168$. in Tables \ref{tabcomplres1} \& \ref{tabcomplres2}. In both tables, we repeat the worst case time complexity of each of the four variations of NICE, along with the CPU times obtained for both the ANN and RF classifiers. Furthermore, we ``fill in'' the complexity functions based on the specific $k$ and $m$ values of the two datasets, which results in an Order of \#``operations''.

In both tables we observe that the Order of \#``operations'' can become quite large (we intentionally chose two datasets with some of the largest $k$ and $m$ values), but that the impact on the CPU times remains limited. E.g., for ``adult'' with the ANN, the CPU times remain below 100 milliseconds. Even with a higher value for $m$ (ANN for ``clean2'') the CPU times are still quite small. The only noticeable increase comes from NICE(\texttt{plaus}), which can be attributed to the AE. For the RF classifier, we notice the CPU times are in general larger than those for the ANN, but are still below 2 seconds, with NICE(\texttt{plaus}) being the only real exception.

In summary, the CPU times remain small, even for some of these larger (in terms of $k$ and $m$ values) datasets. However, the frequent use of an AE for NICE(\texttt{plaus}) can have a negative impact, as can be seen in particular for ``clean2'' in Table \ref{tabcomplres2} (the constant by which $g(x)$ is multiplied is much larger than for ``adult''). Combined with the RF classifier requiring considerable more CPU time than the ANN classifier\footnote{Recall that as part of NICE we only classify an instance $x$ according to the trained model $f$ but do not retrain the model itself. The latter happens offline.}, we conclude that the largest CPU times occur for NICE(\texttt{plaus}) with the RF classifier, which can also be observed from Tables \ref{ANNMainResults} \& \ref{RFMainResults}, but that NICE's CPU times remain within reasonable bounds.

\begin{table}[H]
	\small
	\begin{center}
		\begin{tabular}{lllrr}
			\textbf{Algorithm} &\textbf{Worst case time complexity}& \textbf{Order of \#``operations''} & \textbf{CPU time}& \textbf{CPU time}\\
			& & & \textbf{ANN (ms)}& \textbf{RF (ms)}\\
			\hline
			NICE(\texttt{none})& $O(k\cdot m)$& 547,022& 21.95 & 154.11\\
			NICE(\texttt{spars}) & $O(f(x)\cdot m^2 +k\cdot m)$& $f(x)\cdot$ 196 + 547,022 & 29.29 &1110.52\\
			NICE(\texttt{prox}) & $O(m^3+f(x)\cdot m^2 +k\cdot m)$& $f(x)\cdot$ 196 + 549,766 &32.04&1538.50\\
			NICE(\texttt{plaus}) & $O((f(x)+g(x))\cdot m^2 +k\cdot m)$& $(f(x)+g(x))\cdot$ 196 + 547,022 & 66.61&1776.04\\
		\end{tabular}
	\end{center}
	\caption{Example dataset ``adult'' ($k=39,073$; $m=14$).}
	\label{tabcomplres1}
\end{table} 

\begin{table}[H]
	\small
	\begin{center}
		\begin{tabular}{lllrr}
			\textbf{Algorithm} &\textbf{Worst case time complexity}& \textbf{Order of \#``operations''} & \textbf{CPU time}& \textbf{CPU time}\\
			& & & \textbf{ANN (ms)}& \textbf{RF (ms)}\\
			\hline
			NICE(\texttt{none})& $O(k\cdot m)$& 886,704 & 12.19 &40.03\\
			NICE(\texttt{spars}) & $O(f(x)\cdot m^2 +k\cdot m)$& $f(x)\cdot$ 28,224 + 886,704 & 22.49&446.70\\
			NICE(\texttt{prox}) & $O(m^3+f(x)\cdot m^2 +k\cdot m)$& $f(x)\cdot$ 28,224 + 5,628,336 &18.82&1175.36\\
			NICE(\texttt{plaus}) & $O((f(x)+g(x))\cdot m^2 +k\cdot m)$& $(f(x)+g(x))\cdot$ 28,224 + 886,704 &433.14& 17,359.46\\
		\end{tabular}
	\end{center}
	\caption{Example dataset ``clean2'' ($k=5278$; $m=168$).}
	\label{tabcomplres2}
\end{table} 

\subsection{\label{AppendixParameter}Parameters classification models}

For both classifiers we used the scikit-learn \cite{scikit-learn} implementation which is \\
sklearn.ensemble.RandomForestClassifier for an RF and sklearn.neural\_network.MLPclassifier for an ANN. A five-fold cross-validation grid search is performed with the values of Table \ref{Hyperparameters} where the best performing model is selected based on the ROC AUC score. For the RF the hyperparameter class\_weight is set to 'balanced' and all other hyperparameters are set to default. The ANN always consists of one hidden layer for which the number of neurons in the grid is relative to the size of the input layer (k) with a minimum of 2 neurons. For example for dataset clean2, the number of input neurons is 168, which results in the following grid for the parameter hidden\_layer\_sizes: 2, 25, 50, 76, 101, 126, 151, 176, 202, 227 and 252.

\begin{table}[H]
\begin{tabular}{ll|l}
                                   &                      & \textbf{Parameter Values} \\ \hline
\textbf{RF}  & n\_estimators        & 50, 100, 250, 500, 1000       \\
                                   & max\_depth           & 1, 2, 5, 10, 25, None         \\ \hline
\textbf{ANN} & hidden\_layer\_sizes & 2 to 1.5k, step=0.15k    
\end{tabular}
\caption{Parameter grid used in both cross-validations.}
	\label{Hyperparameters}
\end{table}

\subsection{\label{Appendixmulticlass}Multiclass Reward Functions}

To apply NICE to multiclass, our reward functions need a more general definition. For binary classification we assumed two classes (-1 and 1) for which a classifier $f$  maps $\mathbb{R}^m$ in the class score vector such that $f(x) \in [-1,1]$. For multiclass classification, it is no longer possible to project the scores of our model in such a one-dimensional vector. Therefore, we assume a $m$-dimensional feature space $X \subset \mathbb{R}^m$  consisting of both categorical and numerical features, a feature vector $ x\in X$ with a corresponding label denoted as $y \in Y = \{0,n\}$ and a trained classification model $h$ that maps $\mathbb{R}^m$ in an n-dimensional class probability vector where $h_i(x)$ corresponds to the probability of $x$ belonging to class i.

There are two options to generate multiclass counterfactual explanations \cite{vermeire2022explainable}. First, one might be interested in a counterfactual explanation from a specific class and second, one might be interested in a counterfactual explanation from any class. We propose the following general reward function for both cases.

\begin{equation}
R(x) = \frac{(h_c(x_{i-1,b})-h_o(x_{i-1,b}))-((h_c(x)-h_o(x))}{sparsity(x_{i-1,b},x)}\\
\label{sparsmulticlasslong}
\end{equation}

Equation \ref{sparsmulticlasslong} can be simplified as follow because the sparsity increase every step is equal to 1.
\begin{equation}\label{sparsmulticlass}
R(x) = h_c(x_{i-1,b})-h_0(x_{i-1,b})-h_c(x)+h_0(x)\\
\end{equation}

The definition of $h_c$ and $h_o$ is different depending on the type of counterfactual we are looking for. To find a valid counterfactual from a specific class, the probability of this class has to be higher than the probability of all other classes. In this case the counterfactual probability $h_c$ is equal to the probability of this specific class c, and $h_o$ is the maximum probability of all other class probabilities:

\begin{equation}
h_o(x)= max\{h_i(x): i \in \lbrack0,n \rbrack \sim c\}
\end{equation}

For the second case, where we look for a counterfactual from any class, we want any class probability to be higher than the probability of the original class. In this case we define $h_o$ as the probability of the class for which the original instance to explain had the highest probability and $h_c$ as the maximum probability of all other classes.

\begin{equation}
h_c(x)= max\{h_i(x): i \in \lbrack0,n \rbrack \sim o\}
\end{equation}

The proposed reward function in equation \ref{sparsmulticlass} can be reduced to our reward function \ref{SparsLoss} for binary classification. To do this we have to project the probabilities of both classes into the one dimensional score vector \lbrack -1,1 \rbrack by taking the following assumptions:

\begin{equation}
h_0(x) = \hat{y}\cdot\frac{-f(x)-1}{2} \textrm{  and   } h_c(x) = \hat{y}\cdot\frac{f(x)+1}{2}
\end{equation}

Replacing these in Equation \ref{sparsmulticlass} results in:

\begin{equation}
R(x) = \hat{y}\cdot(\frac{f(x_{i-1,b})+1}{2}-\frac{-f(x_{i-1,b})-1}{2}-\frac{f(x)+1}{2}+\frac{-f(x)-1}{2})\\
\end{equation}

\begin{equation}
R(x) = \hat{y}\cdot(f(x_{i-1,b})-f(x))
\end{equation}

which is equal to our sparsity reward function \ref{SparsLoss}.

\end{document}